\pdfoutput=1

\documentclass[11pt]{article}

\usepackage[]{acl}

\usepackage{times}
\usepackage{latexsym}

\usepackage[T1]{fontenc}

\usepackage[utf8]{inputenc}

\usepackage{microtype}

\usepackage{soul}
\usepackage{url}
\usepackage{graphicx}
\usepackage{amsmath}
\usepackage{amsthm}
\usepackage{xspace}
\usepackage{pifont}%
\usepackage{caption}
\usepackage{subcaption}
\usepackage{arydshln}
\usepackage[ruled,noline,linesnumbered]{algorithm2e}
\usepackage{multirow}
\usepackage{xcolor}
\usepackage{adjustbox}
\usepackage{lipsum} 
\usepackage{booktabs,arydshln}
\usepackage{tabulary}
\usepackage{tabstackengine}
\usepackage{tablefootnote}
\usepackage{enumitem}
\setlist{itemsep=1pt,parsep=1pt,leftmargin=\parindent}
\usepackage{lingmacros}

\newcommand{\bn}{\texttt{\textbackslash n}}
\newcommand{\sent}{\{$x$\}}

\title{Analyzing the Performance of GPT-3.5 and GPT-4 \\ in Grammatical Error Correction}

\author{Steven Coyne$^{1,2}$
\quad \textbf{Keisuke Sakaguchi}$^{1,2}$ \\ 
\textbf{Diana Galvan-Sosa}$^{1,2}$ 
\quad \textbf{Michael Zock}$^{3}$ 
\quad \textbf{Kentaro Inui}$^{1,2}$\\
$^{1}$\text{Tohoku University} \quad $^{2}$\text{RIKEN} \quad $^{3}$\text{LIS, Aix-Marseille University} \\
  \texttt{coyne.steven.charles.q2@dc.tohoku.ac.jp} \\
  \texttt{\{keisuke.sakaguchi,dianags,kentaro.inui\}@tohoku.ac.jp} \\
  \texttt{michael.zock@lis-lab.fr}
}
 
\begin{document}
\maketitle

\begin{abstract}
GPT-3 and GPT-4 models are powerful, achieving high performance on a variety of Natural Language Processing tasks.
However, there is a relative lack of detailed published analysis of their performance on the task of grammatical error correction (GEC). 
To address this, we perform experiments testing the capabilities of a GPT-3.5 model (\texttt{text-davinci-003}) and a GPT-4 model (\texttt{gpt-4-0314}) on major GEC benchmarks. 
We compare the performance of different prompts in both zero-shot and few-shot settings, analyzing intriguing or problematic outputs encountered with different prompt formats. 
We report the performance of our best prompt on the BEA-2019 and JFLEG datasets, finding that the GPT models can perform well in a sentence-level revision setting, with GPT-4 achieving a new high score on the JFLEG benchmark. 
Through human evaluation experiments, we compare the GPT models' corrections to source, human reference, and baseline GEC system sentences and observe differences in editing strategies and how they are scored by human raters.
\end{abstract}

\section{Introduction}

Over the past few years, significant strides have been made in the field of Natural Language Processing (NLP).
OpenAI's GPT models, including GPT-3~\cite{brown2020language} and GPT-4~\cite{openai2023gpt4}, have gained widespread attention among researchers and industry practitioners and demonstrated impressive performance across a variety of tasks in both zero-shot and few-shot settings.

However, information about these models' performance in the task of grammatical error correction (GEC) is still relatively scarce.
OpenAI's technical reports do not include benchmark scores for GEC, as are present for other tasks such as Question Answering.
As OpenAI updates its latest model, there have been only a few studies that try to shed some light on GPT’s performance on the GEC task. 
These works, which we discuss further in Section~\ref{sec:background}, present a preliminary analysis on \texttt{text-davinci-002} and \texttt{gpt-3.5-turbo}. 
Our work seeks to add to and complement these, targeting different GPT models, presenting a more fine-grained prompt and hyperparameter search, and collecting comparative edit quality ratings from human annotators.

\begin{figure}[t!]
\centering
\includegraphics[width=0.9\columnwidth]{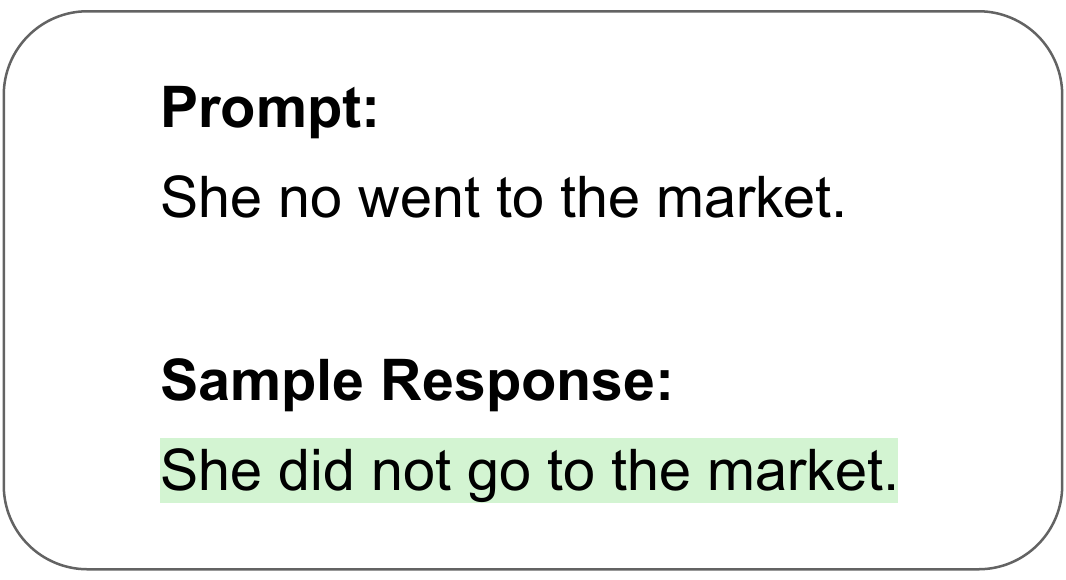}
\caption{OpenAI's example prompt for ``grammar correction,'' showing an input and output (highlighted in green) for the sentence-level revision task.
Our experiments with GPT-3.5 and GPT-4 are based on this pattern.}

\label{fig:intro_example}
\end{figure}

In this work, we assume a prompt setting in which the input is a single potentially ungrammatical sentence and the output is a single correction, as seen in Figure \ref{fig:intro_example}.
We have chosen this setting to match the format of widely used GEC benchmarks which are scored by comparing parallel sentences. 
In addition, we assume a specific task setting of GEC for text revision, taking an ill-formed sentence as input and producing a well-formed version of the sentence which preserves the perceived meaning.

Following a prompt search, we report the performance of \texttt{text-davinci-003}, as well as a current GPT-4 model (\texttt{gpt-4-0314}), on GEC benchmark test sets. 
We then define a subset of sentences and perform side-by-side comparisons of the GPT models' generations, the outputs of two baseline GEC systems, and the human reference edits included in the benchmark datasets.
We report scores from both automated metrics and human raters and perform qualitative analysis of the differences between the respective corrections. 
We also describe our prompt development process and the effect of the temperature hyperparameter on GPT-3.5 and GPT-4's performance on this task.

Based on our experiments, we observe that:
\begin{itemize}
    \item Given a suitable prompt, the GPT models behave reliably in the single-sentence prompt setting, generating no unexpected sequences such as comments or new lines. 
    \item The models show strong performance on the sentence revision task, with GPT-4 achieving a new high score on the JFLEG test set.
    \item The models exhibit some prompt sensitivity. Both the error correction quality and the reliability of the output format differ significantly based on simple changes to wording or punctuation. 
    \item Using our final prompt, the models seem to favor fluency corrections, underperforming on metrics and datasets which rely on a single reference with minimal edits, but performing well on fluency edit tasks and in human evaluations.
    \item The models occasionally over-edit, changing the meaning of a sentence during correction, or expanding fragments with new material.
    \item As a result of the above, different automatic metrics and human raters sometimes disagree on the relative quality of corrections. We examine some cases of this in Section~\ref{Discussion}.
\end{itemize}

Our experimental results emphasize the importance of the specific task setting and choice of benchmark when prompt engineering for large language models such as GPT-3.5 and GPT-4. 

\begin{table*}[ht]
\small
\centering
\begin{tabular}{@{} l p{7.8cm} cccccc @{}}
\toprule[.1em]
       &         & \multicolumn{3}{c}{GPT-3.5} & \multicolumn{3}{c}{GPT-4} \\
No.    &  Prompt & $\tau$=0.1 & $\tau$=0.5 & $\tau$=0.9 & $\tau$=0.1 & $\tau$=0.5 & $\tau$=0.9\\ \midrule[.1em]
1 & Make this sound more fluent:~\bn \bn~\sent & 0.314 & 0.301 & 0.266 & 0.245 & 0.240 & 0.230 \\ \midrule
2 & Update to fix all grammatical and spelling errors:~\bn \bn~\sent & 0.368 & 0.355 & 0.330 & 0.484 & 0.481 & 0.474\\ \midrule
3 & Improve the grammar of this text:~\bn \bn~\sent & 0.494 & 0.486 & 0.459 & 0.427 & 0.421 & 0.414\\ \midrule
4 & Correct this to standard English:~\bn \bn~"\sent" & 0.503 & 0.500 & 0.486 & 0.429 & 0.424 & 0.412\\ \midrule
5 & Act as an editor and fix the issues with this text:~\bn \bn~\sent & 0.516 & 0.505 & 0.494 & 0.444 & 0.444 & 0.435\\ \midrule
6 & Original sentence:~\sent~\bn~Corrected sentence: & 0.552 & 0.547 & 0.533 & 0.523 & 0.521 & 0.520 \\ \midrule
7 & Correct this to standard English:~\bn \bn~\sent & 0.559 & 0.554 & 0.542 & 0.452 & 0.453 & 0.444 \\ \midrule
8 & Correct the following to standard English:~\bn \bn~Sentence:~\sent~\bn~Correction: & 0.569 & 0.564 & 0.551 & 0.495 & 0.488 & 0.480\\ \midrule
9 & Fix the errors in this sentence:~\bn \bn~\sent & 0.569 & 0.566 & 0.554 & 0.541 & 0.542 & 0.534 \\ \midrule
10 & Reply with a corrected version of the input sentence with all grammatical and spelling errors fixed. If there are no errors, reply with a copy of the original sentence.~\bn \bn~Input sentence:~\sent~\bn~Corrected sentence: & \textbf{0.582} & 0.581 & 0.577 & \textbf{0.601} & 0.599 & 0.597 \\
\bottomrule
\end{tabular}
\caption{Performance of different prompts and temperature parameter combinations in a zero-shot GEC setting using GPT-3.5 and GPT-4. All scores are GLEU scores on the JFLEG development set. \sent~represents a source sentence. \bn~ represents a line break. Bold numbers indicate the best-performing combinations.}
\vspace{-0.0cm}
\label{tab:prompts}
\end{table*}

\section{Background}\label{sec:background}

\subsection{OpenAI Models}

Following the success of Transformer-based large language models (LLMs) on several NLP tasks, in which increasing the number of the model's parameters consistently showed improvements,~\citet{brown2020language} trained a 175 billion parameter auto-regressive LM: GPT-3.
GPT-3.5 models are refined from GPT-3 using reinforcement learning from human
feedback~\cite{ouyang2022training}. 
The successor to these, GPT-4, is assumed to be even larger, but the parameter counts were not described in its technical report~\cite{openai2023gpt4}. 
Both models were evaluated on ``over two dozen NLP datasets'', whose tasks range from Question Answering (QA) to Natural Language Inference (NLI) and Reading Comprehension (RC).
GPT-4 was additionally tested on a set of exams that were originally designed for humans
However, no GEC dataset was considered in either of the models' evaluation, necessitating independent task-specific analysis.

\citet{ostlingreally} use a single 2-shot prompt to investigate \texttt{text-davinci-002} in Swedish GEC, finding its performance strong considering it was trained on very little Swedish text.

Following the release of ChatGPT, \citet{wu2023chatgpt} assess its GEC capabilities using a single zero-shot prompt on the CoNLL-2014 dataset \cite{ng-etal-2014-conll}.
\citet{fang2023chatgpt}, investigate \texttt{gpt-3.5-turbo} with both zero-shot and few-shot prompting, as well as human evaluations of the results. 
These studies both find that the GPT models tend to make fluency edits and over-corrections.

Our work differs from the above in the models assessed, the nature of our prompt search, which is more fine-grained in order to investigate prompt sensitivity, and in the aims of our human experiments.
The previous studies on ChatGPT ask participants to identify phenomena such as over-corrections and under-corrections, whereas our experiment elicits comparative error quality ratings.

\subsection{Grammatical Error Correction}

Writing is not an easy task.
Given a goal, we have to decide what to say and how to say it, making sure that the chosen words can be integrated into a coherent whole and conform to the grammar rules of a language \cite{zock2017use}.
This has motivated the NLP community to develop innovative approaches for writing assistance, which are particularly focused on error correction.

GEC research can generally be defined in terms of one of two broad task settings.
The first is education for language learners, in which case easily comprehensible minimal edits are employed, with an emphasis on achieving \textit{grammaticality} but otherwise leaving the sentence as-is.
The other is a revision task in which a sequence is edited to sound \textit{fluent} and natural, and any number or type of changes can be applied as long as the intended meaning, as interpreted by the editor, is preserved.

Research on GEC has primarily been investigated based on the CoNLL-2014 and BEA-2019 \cite{bryant-etal-2019-bea} shared tasks, where systems are evaluated by {F$_{0.5}$} score. 
Since the datasets provided by these two tasks focus on \textit{grammaticality}, \citet{jfleg} released the JFLEG dataset as a new gold standard to evaluate how \textit{fluent} a text is.
Results on this dataset are evaluated with GLEU~\cite{Napoles2015-en}, which relies on n-gram overlap rather than the number of error corrections found in a sentence.

The best systems on each of the aforementioned tasks show a variety of approaches: classification with logistic regression \cite{qorib-etal-2022-frustratingly}, a combination of Statistical and Neural Machine Translation \cite{grundkiewicz-junczys-dowmunt-2018-near, junczys-dowmunt-etal-2018-approaching, kiyono-etal-2019-empirical}, sequence tagging with encoder-only Transformer models \cite{omelianchuk-etal-2020-gector, tarnavskyi-etal-2022-ensembling}, a multilayer CNN encoder-decoder \cite{chollampatt2018multilayer}, and Transformers-based encoder-decoder models \cite{stahlberg-kumar-2021-synthetic, kaneko-etal-2020-encoder}.
\section{Prompt Engineering}

GPT models are autoregressive decoder-only language models with a natural language text prompt as input.
In our task, given an instruction prompt $c$ and input sentence $x$, GPT models generate a text sequence ($y$, tokenized as $(w_1, w_2, \dots w_T)$) based on the following log likelihood:
\begin{align*}
\log p_{\theta}(y|c,x) = \sum_{t=1}^{T} \log p_{\theta}(w_t | c, x, w_{< t-1})
\end{align*}

To best apply the GPT models to this task, it is necessary to first devise an appropriate prompt. 
Therefore, our first step is prompt engineering.

Since the format and even exact wording of a large language model's prompts can have a significant effect on task performance \cite{jiang-etal-2020-know,shin-etal-2020-autoprompt,schick-schutze-2021-just}, we design several different candidate prompts for the GEC task, starting with a zero-shot setting.
Table~\ref{tab:prompts} shows the zero-shot prompts we experimented with, as well as their results. 
Elsewhere in this paper, we will refer to these prompts by number based on their index from this table. 
We begin the prompt search with GPT-3.5 using OpenAI's example prompt for grammatical error correction:\footnote{\url{https://platform.openai.com/examples/default-grammar}, as of April 22, 2023}

\begin{enumerate}
    \item[] Correct this to standard English:\bn\bn
\end{enumerate}

Interestingly, this prompt is defined within the \textsc{completions} endpoint in the OpenAI API.
As an \textsc{edits} endpoint also exists, it may occur to a user to define this task with that endpoint, as grammatical error correction can be considered an editing task.
In our initial experiments, however, we found that the performance of the \textsc{edits} endpoint in this task lagged behind that of the \textsc{completions} endpoint, so we continued our prompt engineering experiments using \textsc{completions} as seen in the example.
Unlike the GPT-3.5 model, GPT-4 only has a \textsc{chat} completion endpoint available via the API.
To maintain similarity across experiments, we submit our prompts to GPT-4 as a single input as the ``user'' role, without defining a system message. \par
We start our prompt engineering experiments with slight modifications to the wording of the example prompt, such as adding quotes to the target sentence, as seen in Prompt \#4.
We then experiment with ``fields'' such as ``Sentence:'' and ``Correction:'', as seen in Prompt \#8.
These relatively small adjustments are designed to test the GPT models' prompt sensitivity.
Finally, we experiment with a more complex prompt, \#10, which specifies a behavior when the sentence is already correct.

In addition, we use nucleus (top-p) sampling~\cite{Holtzman2020The} to generate tokens, repeating experiments with temperature hyperparameters $\tau$ of 0.1, 0.5, and 0.9.\footnote{Other hyperparameters used include logprobs=0, num\_outputs=1, top\_p=1.0, and best\_of=1}

To select the best prompt and temperature combination, we use GLEU scores on the JFLEG development set.

After identifying the best zero-shot prompt, we proceeded to experiments in a few-shot setting, adding one or more example sentence-correction pairs to our best zero-shot prompt to demonstrate the GEC task. 
We experimented with up to six example sentence-correction pairs.
\section{Evaluation Experiments} \label{sec:experiments}

\subsection{Data and Benchmarks}

We use two benchmark datasets: the BEA-2019 shared task dataset and JFLEG. For GEC benchmark scores, we use the test set for both. For qualitative analysis and a human evaluation experiment in which different corrections are compared side-by-side, we define a smaller sample of 200 sentences. We select the first 100 sentences each from BEA-2019 development set\footnote{We use the development set (from the W\&I + LOCNESS dataset~\cite{bryant-etal-2019-bea}) because human-written references are not publicly available for the test set.} and the JFLEG test set, excluding sentences with fewer than 10 tokens, which were mostly greetings or highly fragmentary.

\subsection{Human Evaluation}
\label{sec:human_eval}
In our study, we use the method from~\citet{sakaguchi-van-durme-2018-efficient}, which efficiently elicits scalar annotations as a probability distribution by combining two approaches: direct assessment and online pairwise ranking aggregation.

For the human evaluation task, we asked crowdworkers to compare and score the quality of corrections, with a focus on maintaining the original meaning and ensuring the output is fluent and natural-sounding. 
Participants rated the following five versions of each sentence:~the source sentence (with no corrections), a human-written reference correction (included in the original datasets), the corrections generated by GPT-3.5 and GPT-4 using our best-performing prompt (as seen in Table \ref{tab:final_prompt}), and an output from baseline GEC models for each benchmark (\citet{yasunaga-etal-2021-lm} for BEA-2019 and \citet{liu-etal-2021-neural} for JFLEG). These systems were chosen due to the availability of their outputs, allowing for direct side-by-side comparisons.

For each comparison, we assign three crowdworkers to score the quality of corrections on a scale of 0 (very poor correction) to 10 (excellent correction).
Additional details about the human evaluation task can be found in the appendix.
\section{Results} \label{sec:results}

\subsection{Prompt Engineering}

Scores for different zero-shot prompts can be seen in Table \ref{tab:prompts}. Consistent with expectations, we find that the content of the prompt is very significant for performance. 
Our best zero-shot prompt has more than double the score of the worst on automated metrics. 
It is also clear that the temperature hyperparameter has an effect on performance in this task, with lower temperatures consistently performing better than higher temperatures. 

\begin{table}[t]
\begin{adjustbox}{max width=\columnwidth}
\centering
\small
\begin{tabular}{@{} l| c c c c c c @{}}
\toprule[.1em]
\#-shot & 1 & 2 & 3 & 4 & 5 & 6 \\ \midrule
GPT-3.5 & 0.587 & \textbf{0.590} & 0.585 & 0.584 & 0.586 & 0.584 \\
\midrule
GPT-4 & 0.599 & \textbf{0.600} & 0.594 & 0.593 & 0.593 & 0.588 \\
\bottomrule
\end{tabular}
\end{adjustbox}
\caption{Few-shot performance of Prompt \#10 with a variable number of example sentence-correction pairs. All scores are GLEU scores on the JFLEG dev set.}
\label{tab:num_prompts}
\end{table}
\begin{table*}[t]
\small
\centering
\begin{tabular}{p{0.8\linewidth}}
\toprule[.1em]
Reply with a corrected version of the input sentence with all grammatical and spelling errors fixed. If there are no errors, reply with a copy of the original sentence.\\
\\
Input sentence: I think smoke should to be ban in all restarants.\\
Corrected sentence: I think smoking should be banned at all restaurants.\\
\\
Input sentence: We discussed about the issu.\\
Corrected sentence: We discussed the issue.\\
\\
Input sentence: \sent\\
Corrected sentence: \\
\bottomrule[.1em]
\end{tabular}
\caption{\label{tab:final_prompt}
The final prompt chosen following our prompt engineering experiments. All GPT-3.5 and GPT-4 scores and examples in Sections~\ref{sec:results} and \ref{Discussion} use outputs made with this prompt.}
\end{table*}
Moving on to few-shot prompts, we experimented by adding examples to Prompt \#10. 
Results from this experiment can be seen in Table \ref{tab:num_prompts}. 
We find that for GPT-3.5, performance modestly improves over the zero-shot prompt in all cases, but peaks at two examples. 
For GPT-4, the few-shot examples seem to have a negligible or slight negative effect, with two examples once again scoring the highest among few-shot prompts.

Against expectations, there were many zero-shot prompts in which GPT-3.5 outperformed GPT-4.
This was observed in all prompts except for Prompts \#2 and \#10, in which GPT-4 performed better.
However, several of the same trends can be seen for both models, such as the tendency for lower temperature to improve performance.
Ultimately, both models showed the strongest performance on the same prompt, and had the same best number of examples in few-shot experimentation.

Observing the outputs during zero-shot prompt development, we identified several interesting phenomena and issues, which we will describe here.

\paragraph{``Meta'' information:} Sometimes, the models include a note that no edits were made (\ref{ex:inout1}), declare that a sentence is correct (\ref{ex:inout2}), or respond in a chat-like manner (\ref{ex:inout3}). Examples can be found below:

\enumsentence{
\small
\textbf{Input:} \\
People also do not do nothing.

\textbf{Output (Prompt \#5):} \\
People also do not do nothing. (no change)
\label{ex:inout1}
}

\enumsentence{
\small
\textbf{Input:} \\
They try new things.

\textbf{Output (Prompt \#9):} \\
There are no errors in this sentence.
\label{ex:inout2}
}

\enumsentence{
\small
\textbf{Input:} \\
But ideas of how you can solve this problem will actually give you a solution to the problem. 

\textbf{Output (Prompt \#3):} \\
By revising the sentence, it can be improved to: Generating ideas for how to solve this problem will help you find a solution.
\label{ex:inout3}
}

\paragraph{Outputs with extra punctuation:}
Another interesting phenomenon occurs when a source sentence does not end with proper punctuation. The models may start by adding the appropriate punctuation mark, then continue by correcting the sentence, resulting in an extra punctuation mark at the beginning of the output. 
An example is seen below:

\enumsentence{
\small
\textbf{Input:} \\
The second one is to specialize in one specific subject, one has his own tast in studing, so let us talk about this subject

\textbf{Output (Prompt \#2):} \\
. The second one is to specialize in one specific subject; one has their own taste in studying, so let us talk about this subject.
\label{ex:inout4}
}

\paragraph{Outputs with spurious expansions:}
A more severe version of the aforementioned issue, in which the model continues generating text based on the context of the source sentence. 
This can result in spurious expansions of the original writing, such as repetitions at the beginning of the sequence or extra lines. 
An example is seen below:

\enumsentence{
\small
\textbf{Input:} \\
For example, if the city says that 8am~9 am is the most busiest time, than worker would argues that they drive a bit later 

\textbf{Output (Prompt \#7):} \\
 when the traffic is more light. \\
For example, if the city says that 8am-9am is the busiest time, then workers would argue that they should drive a bit later when the traffic is lighter.
\label{ex:inout5}
}
%
\begin{table*}[ht]
\small
\centering
\begin{tabular}{@{} l c c  @{}}
\toprule[.1em]

\multicolumn{1}{c}{}
&
\multicolumn{1}{c}{\textbf{BEA-2019 (Test)}}
&
\multicolumn{1}{c}{\textbf{JFLEG (Test)}}
\\
             & {F$_{0.5}$}  & {GLEU}  \\  \midrule
Source (Uncorrected)       & 0 & 40.54 \\
Human Reference    & - & 62.37 \\ \midrule
GECToR+BIFI \cite{yasunaga-etal-2021-lm}     & \textbf{72.9} & - \\
ELECTRA-VERNet \cite{liu-etal-2021-neural}  & 67.28 & 61.61 \\ \midrule
``GPT-3'' \cite{yasunaga-etal-2021-lm}     & 47.6 & - \\
GPT-3 (text-davinci-001) \cite{schick2022peer}    & - & 60.0 \\
GPT-3.5 (text-davinci-003)     & 49.66 & 63.40 \\
GPT-4 (gpt-4-0314)       & 52.79 & \textbf{65.02} \\ \bottomrule
\end{tabular}
\caption{GEC Benchmark scores for GPT-3.5 and GPT-4 using our final prompt, alongside those of baseline GEC systems and previously reported scores for GPT-3. The best scores are in bold.}
\vspace{-0.0cm}
\label{tab:benchmark_results}
\end{table*}

In this case, the added text and newline at the beginning are problematic, resulting in an issue in the GLEU evaluation script by breaking the symmetry of lines in the input files. 
It is also not desirable to show this as-is to a user of a GEC system, since the output is noticeably strange.

For our final prompt, we choose Prompt \#10 with two examples, which can be seen in Table~\ref{tab:final_prompt},
Despite GPT-4's slightly higher performance with a zero-shot prompt, we use this 2-shot prompt with both models in our experiments in order to observe the differences between the models given the exact same input sequence.
This ``best'' prompt produced few or none of the above unexpected outputs with either GPT-3.5 or GPT-4. 
There were no repetitions or new lines.
This emphasizes the importance of prompt design when applying GPT models.

\subsection{Benchmark Scoring} 

GEC benchmark scores, calculated on the BEA-2019 and JFLEG test sets, are shown in Table~\ref{tab:benchmark_results}. 
To score GPT-3.5 and GPT-4's outputs against the references and baseline systems, we use the standard scores for each dataset, {F$_{0.5}$} for BEA-2019 and GLEU for JFLEG.
When interpreting results, note that in the BEA-2019 benchmark, the {F$_{0.5}$} score is essentially 0 for the source. 
The human reference score is unknown, as the reference edits are part of the withheld test set.
To obtain the score for the ``Human Reference'' corrections in the JFLEG dataset, which has multiple references, we randomly selected one human reference file and compared with it the other three references.

The results show that the GPT models perform well on the JFLEG test set, with GPT-4 obtaining a score that is the highest yet reported to the best of our knowledge. 
In contrast, the scores on the BEA-2019 test set are well below those of the baseline systems.
We discuss this disparity in Section~\ref{Discussion}.

\subsection{Human Evaluation and Subset Analysis} \label{section:results}
\begin{table*}[ht]
\small
\centering
\begin{tabular}{@{} l r c c c c c @{}}
\toprule[.1em]

\multicolumn{1}{c}{}
&
\multicolumn{3}{c}{\textbf{BEA-2019 (Dev Subset)}}
&
\multicolumn{3}{c}{\textbf{JFLEG (Test Subset)}}
\\
             & {F$_{0.5}$}  & {Human} & {Scribendi} & {GLEU} & {Human} & {Scribendi}\\ 
             Scale: & (0-100)  & (0-1)  & (0-1) & (0-100)  & (0-1) & (0-1) \\  \midrule
Source       & 0 & 0.449 & 0 & 36.51 & 0.465 & 0\\
Reference    & \textbf{83.97} & 0.706 & \textbf{0.83} & 54.63 & 0.712 & 0.74\\
Baseline     & 39.14  & 0.568 & 0.67 & 57.70 & 0.662 & 0.71\\
GPT-3.5      & 37.87 & 0.769 & 0.71 & 63.02 & \textbf{0.819} & \textbf{0.78}\\
GPT-4        & 37.99 & \textbf{0.788} & 0.75 & \textbf{63.78} & 0.809 & 0.75\\ \bottomrule
\end{tabular}
\caption{Comparison of automated metrics and human evaluation scores for different versions of sentences in our human evaluation subset of 100 sentences from each dataset, as described in section~\ref{sec:human_eval}. The best scores are in bold. In human evaluations, the difference between GPT-3.5 and GPT-4 is not statistically significant for either the BEA-2019 or JFLEG benchmarks ($p>$0.19, $p>$0.4).}
\vspace{-0.0cm}
\label{tab:subset_results}
\end{table*}

For the subset of 100 sentences each from the BEA-2019 development set and the JFLEG test set, we gather human ratings as described in Section~\ref{sec:human_eval} and place them alongside the respective datasets' automated metrics.
Additionally, we apply a ``reference-less'' automatic metric, Scribendi Score~\cite{islam-magnani-2021-end}, which assesses grammaticality, fluency, and  syntactic similarity using token sort ratio, levenshtein distance ratio, and perplexity as calculated by GPT-2. 
We use an unofficial implementation,\footnote{\url{https://github.com/gotutiyan/scribendi_score}} as the authors seem not to have made their code available. 

The scores from our experiments are shown in Table \ref{tab:subset_results}.
Note that the BEA-2019 benchmark's {F$_{0.5}$} score for human reference is not 100 despite the same single reference because the edits are automatically extracted in the evaluation script~\cite{bryant-etal-2019-bea}. 
Scores from Scribendi are returned on a per-sentence basis, so we report the mean for each output file. A score of 0 indicates no edits.

The results suggest that GPT-3.5 and GPT-4 achieve high performance on the task of GEC according to human evaluations and the automatic metrics, with a majority of the best scores being obtained by either GPT-3.5 or GPT-4. 
\section{Discussion} \label{Discussion}

\subsection{Scoring Disparities} 

The results in Tables~\ref{tab:benchmark_results} and \ref{tab:subset_results} show that GPT-3.5 and GPT-4 achieve strong performance on the sentence revision task as measured by GLEU score on the JFLEG dataset, human ratings, and Scribendi scores, outperforming the baseline systems on these metrics.
However, their {F$_{0.5}$} scores on the BEA-2019 datasets are comparatively lower.

We believe that this is a result of differences in the priorities expressed in the human reference edits present in the two datasets.
In the BEA-2019 dataset, there is a single reference for each sentence, generally with what could be described as minimal edits. 
Meanwhile, our primary task setting is one of sentence revision, and our prompt engineering experiments were performed using JFLEG, a benchmark for fluency.
This seems to have contributed to a propensity for the GPT models to output fluency corrections which display extensive editing.
These are scored well on JFLEG's GLEU metric, but penalized on BEA-2019's {F$_{0.5}$} metric.

This is supported by the fact that the models were given similar scores in both datasets by human raters and the Scribendi metric, which is not connected to references from either dataset and is thus not affected by any differences between the reference edits found in BEA-2019 and JFLEG.

\subsection{Qualitative Analysis}

The scores discussed above describe the performance of the different systems in aggregate.
However, there are a number of cases in which the GPT models' outputs are given scores which differ from those assigned by the automated metrics.
Additionally, there are cases in the human evaluation experiments in which the GPT models significantly over-perform or under-perform the human reference edits or the baseline systems. 
We consider a performance discrepancy notable if the candidate sentences show a difference of more than 2 points in the mean of human ratings assigned to them.

To investigate such cases and better understand the behavior of the GPT models as grammatical correction systems, we examine the models' outputs in parallel with the source and reference sentences and those of the baseline error correction systems. Below, we present output sentences along with their respective scores from human raters.

\paragraph{GPT Models Outscoring Human References}

We found 24 cases in JFLEG Test and 14 cases in BEA-2019 Dev in which the GPT models both outscored the human reference edits. 
We find that these cases usually occur when a human editor leaves a grammatical error or non-fluent construction unchanged, but the GPT models revise it. An example can be seen below:

\enumsentence{
\small
\textbf{Source Sentence: (3)} \\
This reminds me of a trip that I have recently been to and the place is Agra.

\textbf{Human Reference: (3.66)} \\
This reminds me of a trip that I have recently been on and the place I visited was Agra.

\textbf{Baseline System: (3)} \\
This reminds me of a trip that I have recently been to and the place is Agra.

\textbf{GPT-3.5: (9.66)} \\
This reminds me of a trip I recently took to Agra.

\textbf{GPT-4: (10)} \\
This reminds me of a recent trip I took to Agra.
\label{ex:inout7}
}

In this case, the edits made by GPT models are the most natural and correct sentences, and are given the highest scores by the raters.
However, this is not to say that the human reference edit was mistaken or inferior, especially if we consider that this example is taken from the BEA dataset, in which minimal edits are common.
Nevertheless, there are also a number of such cases in our subset from JFLEG, where the goal of the task is fluency editing.
This demonstrates that humans tasked with performing or evaluating corrections do not always agree on the ideal extent of revision.

\paragraph{Over-editing} There are some cases in which the GPT models add or change words in a way that results in changes in the meaning of the sentence. An example can be seen below:

\enumsentence{
\small
\textbf{Source Sentence: (4)} \\
I consider that is more convenient to drive a car because you carry on more things in your own car than travelling \underline{by car}.

\textbf{Human Reference: (4)} \\
I consider it more convenient to drive a car, because you carry more things in your own car than when travelling \underline{by car}.

\textbf{Baseline System: (6.67)} \\
I consider that it is more convenient to drive a car because you carry on more things in your own car than travelling \underline{by car}.

\textbf{GPT-3.5: (7.67)} \\
I consider it more convenient to drive a car because you can carry more things in your own car than when travelling \underline{by public transport}.

\textbf{GPT-4: (9)} \\
I consider it more convenient to drive a car because you can carry more things in your own car than when traveling \underline{by public transportation}.	
\label{ex:inout8}
}

Here, it seems likely that public transportation is what the writer is comparing cars to, but the term does not appear in the source sentence.

While such cases in our data generally result in sequences that seem likely, it may be desirable to control for this behavior depending on the GEC task setting.

There are also cases where a fragmentary sentence is expanded by the GPT models.
For these as well, suggesting completions is not necessarily in the scope of GEC.
An example can be seen below:

\enumsentence{
\small
\textbf{Source Sentence: (1.33)} \\
If the film doesn't arrive on time, it immediately.

\textbf{Human Reference: (1.33)} \\
If the film doesn't arrive on time, it immediately.

\textbf{Baseline System: (1.66)} \\
If the film doesn't arrive on time, it will immediately.

\textbf{GPT-3.5: (9.66)} \\
If the film doesn't arrive on time, it will be \underline{cancelled} immediately.

\textbf{GPT-4: (4)} \\
If the film doesn't arrive on time, it will be \underline{shown} immediately.	
\label{ex:inout9}
}

In this case, it seems as if the GPT models, given only this fragment as context, attempt to fix it by adding some plausible verb, with GPT-3.5's completion being judged more reasonable.
However, depending on the task setting, it may be desirable to take some action other than suggesting a correction in these cases. 
For example, a system may simply highlight the sentence as ungrammatical, or perhaps a feedback comment about missing verbs or fragments could be generated instead.
These actions exceed the scope of our experiments, but could certainly be achieved with a more complex writing assistance program.
Whether any such alternative behaviors could reliably be achieved by prompting the GPT models is left to future work.

\paragraph{GPT Models Underperforming}

In the majority of cases in the subset, the GPT models had comparable or superior performance to the baseline systems.
However, there were some cases (4 in BEA-2019 and 7 in JFLEG) where the baseline systems outperformed the GPT models.

The human references were more likely to outperform the GPT models, with 13 cases in BEA-2019 and 10 in JFLEG.
We examine a case of GPT underperformance below:

\enumsentence{
\small
\textbf{Source Sentence: (3.33)} \\
\underline{By the time up} everyone should be gathered up in a certain place.

\textbf{Human Reference: (9.33)} \\
\underline{When the time is up}, everyone should be gathered in a certain place.

\textbf{Baseline System: (6.66)} \\
\underline{By the time everyone gets up}, everyone should be gathered up in a certain place.

\textbf{GPT-3.5: (6.66)} \\
\underline{By the time}, everyone should be gathered in a certain place.

\textbf{GPT-4: (3.33)} \\
\underline{By the time up}, everyone should be gathered in a certain place.
\label{ex:inout6}
}

In this case, only the human editor successfully infers the intended phrase, as judged by the raters.
The baseline edit presents an alternative, grammatically correct possibility.
Meanwhile, the GPT models leave an ungrammatical span of the original sentence unchanged.

This ``under-editing'' behavior is interesting given that we also observe that the GPT models make frequent and extensive edits.
Given the size of our subset, is difficult to generalize about the circumstances in which the models under-edit or over-edit, or if there are ways to control either behavior. 
We leave such investigation to future work.

\section{Conclusion}
We find that the GPT-3.5 and GPT-4 models demonstrate strong performance in grammatical error correction as defined in a sentence revision task.
During prompt and hyperparameter search, we observe that a low temperature hyperparameter is consistently associated with better performance in this task. 
While the models are subject to some prompt sensitivity, our best prompt consistently results in the desired format and behavior.
Our GEC task setting and prompt search resulted in a tendency for the models to produce fluency corrections and occasional over-editing, resulting in high scores on fluency metrics and human evaluation, but comparatively lower scores on the BEA-2019 dataset, which favors minimal edits.

Our experiments emphasize that GEC is a challenging subfield of NLP with a number of distinct subtasks and variables.
Even humans can have conflicting definitions of desirable corrections to ill-formed text, and this may change depending on contexts such as the task setting (e.g. language education, revising an academic paper) and the roles of the editor and recipient (e.g., student and instructor).
It is important to define these variables as clearly as possible in all discussions of GEC.
\section{Limitations}

The scores presented in this paper are based on proprietary models accessed via API. 
They may be updated internally or deprecated in the future.

As this is a preliminary exploration of the behavior of GPT-3.5 and GPT-4 in this task, we limit our experiments to the ten listed prompts, making no claims of an exhaustive search.
We do not try such techniques as chain-of-thought prompting.
We leave such experiments to future research.

For time and budget reasons, the metric scores reported are for a single output file for each dataset and model combination.
Our human annotation experiment was similarly limited by budget, and qualitative analysis was only performed on two hundred sets of candidate sentences.

Due to our sentence revision setting, our experiments focused more on fluency edits than minimal edits, and our human raters tended to prefer the extensive rewrites and that the GPT models often output. 
However, more constrained corrections may be desirable in different GEC task settings, such as in language education, where a learner may more clearly understand the information presented by a minimal edit.
A similar study can be done to investigate how well GPT models can adhere to a minimal editing task.
We leave this to future work.
\section*{Acknowledgments}

This work was supported by JSPS KAKENHI Grant Numbers JP22H00524 and JP21K21343.
Additionally, we would like to thank the authors of the baseline GEC systems for making their model outputs available for comparison.

\bibliography{custom}

\begin{thebibliography}{29}
\expandafter\ifx\csname natexlab\endcsname\relax\def\natexlab#1{#1}\fi

\bibitem[{Brown et~al.(2020)Brown, Mann, Ryder, Subbiah, Kaplan, Dhariwal,
  Neelakantan, Shyam, Sastry, Askell et~al.}]{brown2020language}
Tom Brown, Benjamin Mann, Nick Ryder, Melanie Subbiah, Jared~D Kaplan, Prafulla
  Dhariwal, Arvind Neelakantan, Pranav Shyam, Girish Sastry, Amanda Askell,
  et~al. 2020.
\newblock \href
  {https://proceedings.neurips.cc/paper_files/paper/2020/file/1457c0d6bfcb4967418bfb8ac142f64a-Paper.pdf}
  {Language models are few-shot learners}.
\newblock \emph{Advances in neural information processing systems},
  33:1877--1901.

\bibitem[{Bryant et~al.(2019)Bryant, Felice, Andersen, and
  Briscoe}]{bryant-etal-2019-bea}
Christopher Bryant, Mariano Felice, {\O}istein~E. Andersen, and Ted Briscoe.
  2019.
\newblock \href {https://aclanthology.org/W19-4406} {The {BEA}-2019 shared task
  on grammatical error correction}.
\newblock In \emph{Proceedings of the Fourteenth Workshop on Innovative Use of
  NLP for Building Educational Applications}, pages 52--75, Florence, Italy.
  Association for Computational Linguistics.

\bibitem[{Chollampatt and Ng(2018)}]{chollampatt2018multilayer}
Shamil Chollampatt and Hwee~Tou Ng. 2018.
\newblock A multilayer convolutional encoder-decoder neural network for
  grammatical error correction.
\newblock In \emph{AAAI Conference on Artificial Intelligence}.

\bibitem[{Fang et~al.(2023)Fang, Yang, Lan, Wong, Hu, Chao, and
  Zhang}]{fang2023chatgpt}
Tao Fang, Shu Yang, Kaixin Lan, Derek~F. Wong, Jinpeng Hu, Lidia~S. Chao, and
  Yue Zhang. 2023.
\newblock \href {http://arxiv.org/abs/2304.01746} {Is chatgpt a highly fluent
  grammatical error correction system? a comprehensive evaluation}.

\bibitem[{Grundkiewicz and
  Junczys-Dowmunt(2018)}]{grundkiewicz-junczys-dowmunt-2018-near}
Roman Grundkiewicz and Marcin Junczys-Dowmunt. 2018.
\newblock \href {https://doi.org/10.18653/v1/N18-2046} {Near human-level
  performance in grammatical error correction with hybrid machine translation}.
\newblock In \emph{Proceedings of the 2018 Conference of the North {A}merican
  Chapter of the Association for Computational Linguistics: Human Language
  Technologies, Volume 2 (Short Papers)}, pages 284--290, New Orleans,
  Louisiana. Association for Computational Linguistics.

\bibitem[{Holtzman et~al.(2020)Holtzman, Buys, Du, Forbes, and
  Choi}]{Holtzman2020The}
Ari Holtzman, Jan Buys, Li~Du, Maxwell Forbes, and Yejin Choi. 2020.
\newblock \href {https://openreview.net/forum?id=rygGQyrFvH} {The curious case
  of neural text degeneration}.
\newblock In \emph{International Conference on Learning Representations}.

\bibitem[{Islam and Magnani(2021)}]{islam-magnani-2021-end}
Md~Asadul Islam and Enrico Magnani. 2021.
\newblock \href {https://doi.org/10.18653/v1/2021.emnlp-main.239} {Is this the
  end of the gold standard? a straightforward reference-less grammatical error
  correction metric}.
\newblock In \emph{Proceedings of the 2021 Conference on Empirical Methods in
  Natural Language Processing}, pages 3009--3015, Online and Punta Cana,
  Dominican Republic. Association for Computational Linguistics.

\bibitem[{Jiang et~al.(2020)Jiang, Xu, Araki, and
  Neubig}]{jiang-etal-2020-know}
Zhengbao Jiang, Frank~F. Xu, Jun Araki, and Graham Neubig. 2020.
\newblock \href {https://aclanthology.org/2020.tacl-1.28} {How can we know what
  language models know?}
\newblock \emph{Transactions of the Association for Computational Linguistics},
  8:423--438.

\bibitem[{Junczys-Dowmunt et~al.(2018)Junczys-Dowmunt, Grundkiewicz, Guha, and
  Heafield}]{junczys-dowmunt-etal-2018-approaching}
Marcin Junczys-Dowmunt, Roman Grundkiewicz, Shubha Guha, and Kenneth Heafield.
  2018.
\newblock \href {https://doi.org/10.18653/v1/N18-1055} {Approaching neural
  grammatical error correction as a low-resource machine translation task}.
\newblock In \emph{Proceedings of the 2018 Conference of the North {A}merican
  Chapter of the Association for Computational Linguistics: Human Language
  Technologies, Volume 1 (Long Papers)}, pages 595--606, New Orleans,
  Louisiana. Association for Computational Linguistics.

\bibitem[{Kaneko et~al.(2020)Kaneko, Mita, Kiyono, Suzuki, and
  Inui}]{kaneko-etal-2020-encoder}
Masahiro Kaneko, Masato Mita, Shun Kiyono, Jun Suzuki, and Kentaro Inui. 2020.
\newblock \href {https://doi.org/10.18653/v1/2020.acl-main.391}
  {Encoder-decoder models can benefit from pre-trained masked language models
  in grammatical error correction}.
\newblock In \emph{Proceedings of the 58th Annual Meeting of the Association
  for Computational Linguistics}, pages 4248--4254, Online. Association for
  Computational Linguistics.

\bibitem[{Kiyono et~al.(2019)Kiyono, Suzuki, Mita, Mizumoto, and
  Inui}]{kiyono-etal-2019-empirical}
Shun Kiyono, Jun Suzuki, Masato Mita, Tomoya Mizumoto, and Kentaro Inui. 2019.
\newblock \href {https://doi.org/10.18653/v1/D19-1119} {An empirical study of
  incorporating pseudo data into grammatical error correction}.
\newblock In \emph{Proceedings of the 2019 Conference on Empirical Methods in
  Natural Language Processing and the 9th International Joint Conference on
  Natural Language Processing (EMNLP-IJCNLP)}, pages 1236--1242, Hong Kong,
  China. Association for Computational Linguistics.

\bibitem[{Liu et~al.(2021)Liu, Yi, Sun, Yang, and Chua}]{liu-etal-2021-neural}
Zhenghao Liu, Xiaoyuan Yi, Maosong Sun, Liner Yang, and Tat-Seng Chua. 2021.
\newblock \href {https://aclanthology.org/2021.naacl-main.429} {Neural quality
  estimation with multiple hypotheses for grammatical error correction}.
\newblock In \emph{Proceedings of the 2021 Conference of the North American
  Chapter of the Association for Computational Linguistics: Human Language
  Technologies}, pages 5441--5452, Online. Association for Computational
  Linguistics.

\bibitem[{Napoles et~al.(2015)Napoles, Sakaguchi, Post, and
  Tetreault}]{Napoles2015-en}
Courtney Napoles, Keisuke Sakaguchi, Matt Post, and Joel Tetreault. 2015.
\newblock \href {https://aclanthology.org/P15-2097} {Ground truth for
  grammatical error correction metrics}.
\newblock In \emph{Proceedings of the 53rd Annual Meeting of the Association
  for Computational Linguistics and the 7th International Joint Conference on
  Natural Language Processing (Volume 2: Short Papers)}, pages 588--593,
  Beijing, China. Association for Computational Linguistics.

\bibitem[{Napoles et~al.(2017)Napoles, Sakaguchi, and Tetreault}]{jfleg}
Courtney Napoles, Keisuke Sakaguchi, and Joel Tetreault. 2017.
\newblock \href {https://aclanthology.org/E17-2037} {{JFLEG}: A fluency corpus
  and benchmark for grammatical error correction}.
\newblock In \emph{Proceedings of the 15th Conference of the {E}uropean Chapter
  of the Association for Computational Linguistics: Volume 2, Short Papers},
  pages 229--234, Valencia, Spain. Association for Computational Linguistics.

\bibitem[{Ng et~al.(2014)Ng, Wu, Briscoe, Hadiwinoto, Susanto, and
  Bryant}]{ng-etal-2014-conll}
Hwee~Tou Ng, Siew~Mei Wu, Ted Briscoe, Christian Hadiwinoto, Raymond~Hendy
  Susanto, and Christopher Bryant. 2014.
\newblock \href {https://doi.org/10.3115/v1/W14-1701} {The {C}o{NLL}-2014
  shared task on grammatical error correction}.
\newblock In \emph{Proceedings of the Eighteenth Conference on Computational
  Natural Language Learning: Shared Task}, pages 1--14, Baltimore, Maryland.
  Association for Computational Linguistics.

\bibitem[{Omelianchuk et~al.(2020)Omelianchuk, Atrasevych, Chernodub, and
  Skurzhanskyi}]{omelianchuk-etal-2020-gector}
Kostiantyn Omelianchuk, Vitaliy Atrasevych, Artem Chernodub, and Oleksandr
  Skurzhanskyi. 2020.
\newblock \href {https://doi.org/10.18653/v1/2020.bea-1.16} {{GECT}o{R} {--}
  grammatical error correction: Tag, not rewrite}.
\newblock In \emph{Proceedings of the Fifteenth Workshop on Innovative Use of
  NLP for Building Educational Applications}, pages 163--170, Seattle, WA, USA
  → Online. Association for Computational Linguistics.

\bibitem[{OpenAI(2023)}]{openai2023gpt4}
OpenAI. 2023.
\newblock \href {http://arxiv.org/abs/2303.08774} {Gpt-4 technical report}.

\bibitem[{Ostling and Kurfal{\i}(2022)}]{ostlingreally}
Robert Ostling and Murathan Kurfal{\i}. 2022.
\newblock Really good grammatical error correction, and how to evaluate it.
\newblock \emph{Swedish Language Technology Conference (SLTC)}.

\bibitem[{Ouyang et~al.(2022)Ouyang, Wu, Jiang, Almeida, Wainwright, Mishkin,
  Zhang, Agarwal, Slama, Ray, Schulman, Hilton, Kelton, Miller, Simens, Askell,
  Welinder, Christiano, Leike, and Lowe}]{ouyang2022training}
Long Ouyang, Jeff Wu, Xu~Jiang, Diogo Almeida, Carroll~L. Wainwright, Pamela
  Mishkin, Chong Zhang, Sandhini Agarwal, Katarina Slama, Alex Ray, John
  Schulman, Jacob Hilton, Fraser Kelton, Luke Miller, Maddie Simens, Amanda
  Askell, Peter Welinder, Paul Christiano, Jan Leike, and Ryan Lowe. 2022.
\newblock \href {http://arxiv.org/abs/2203.02155} {Training language models to
  follow instructions with human feedback}.

\bibitem[{Qorib et~al.(2022)Qorib, Na, and Ng}]{qorib-etal-2022-frustratingly}
Muhammad Qorib, Seung-Hoon Na, and Hwee~Tou Ng. 2022.
\newblock \href {https://doi.org/10.18653/v1/2022.naacl-main.143}
  {Frustratingly easy system combination for grammatical error correction}.
\newblock In \emph{Proceedings of the 2022 Conference of the North American
  Chapter of the Association for Computational Linguistics: Human Language
  Technologies}, pages 1964--1974, Seattle, United States. Association for
  Computational Linguistics.

\bibitem[{Sakaguchi and Van~Durme(2018)}]{sakaguchi-van-durme-2018-efficient}
Keisuke Sakaguchi and Benjamin Van~Durme. 2018.
\newblock \href {https://aclanthology.org/P18-1020} {Efficient online scalar
  annotation with bounded support}.
\newblock In \emph{Proceedings of the 56th Annual Meeting of the Association
  for Computational Linguistics (Volume 1: Long Papers)}, pages 208--218,
  Melbourne, Australia. Association for Computational Linguistics.

\bibitem[{Schick et~al.(2022)Schick, Dwivedi-Yu, Jiang, Petroni, Lewis,
  Izacard, You, Nalmpantis, Grave, and Riedel}]{schick2022peer}
Timo Schick, Jane Dwivedi-Yu, Zhengbao Jiang, Fabio Petroni, Patrick Lewis,
  Gautier Izacard, Qingfei You, Christoforos Nalmpantis, Edouard Grave, and
  Sebastian Riedel. 2022.
\newblock \href {http://arxiv.org/abs/2208.11663} {{PEER: A Collaborative
  Language Model}}.

\bibitem[{Schick and Sch{\"u}tze(2021)}]{schick-schutze-2021-just}
Timo Schick and Hinrich Sch{\"u}tze. 2021.
\newblock \href {https://aclanthology.org/2021.naacl-main.185} {It{'}s not just
  size that matters: Small language models are also few-shot learners}.
\newblock In \emph{Proceedings of the 2021 Conference of the North American
  Chapter of the Association for Computational Linguistics: Human Language
  Technologies}, pages 2339--2352, Online. Association for Computational
  Linguistics.

\bibitem[{Shin et~al.(2020)Shin, Razeghi, Logan~IV, Wallace, and
  Singh}]{shin-etal-2020-autoprompt}
Taylor Shin, Yasaman Razeghi, Robert~L. Logan~IV, Eric Wallace, and Sameer
  Singh. 2020.
\newblock \href {https://aclanthology.org/2020.emnlp-main.346} {{A}uto{P}rompt:
  {E}liciting {K}nowledge from {L}anguage {M}odels with {A}utomatically
  {G}enerated {P}rompts}.
\newblock In \emph{Proceedings of the 2020 Conference on Empirical Methods in
  Natural Language Processing (EMNLP)}, pages 4222--4235, Online. Association
  for Computational Linguistics.

\bibitem[{Stahlberg and Kumar(2021)}]{stahlberg-kumar-2021-synthetic}
Felix Stahlberg and Shankar Kumar. 2021.
\newblock \href {https://aclanthology.org/2021.bea-1.4} {Synthetic data
  generation for grammatical error correction with tagged corruption models}.
\newblock In \emph{Proceedings of the 16th Workshop on Innovative Use of NLP
  for Building Educational Applications}, pages 37--47, Online. Association for
  Computational Linguistics.

\bibitem[{Tarnavskyi et~al.(2022)Tarnavskyi, Chernodub, and
  Omelianchuk}]{tarnavskyi-etal-2022-ensembling}
Maksym Tarnavskyi, Artem Chernodub, and Kostiantyn Omelianchuk. 2022.
\newblock \href {https://doi.org/10.18653/v1/2022.acl-long.266} {Ensembling and
  knowledge distilling of large sequence taggers for grammatical error
  correction}.
\newblock In \emph{Proceedings of the 60th Annual Meeting of the Association
  for Computational Linguistics (Volume 1: Long Papers)}, pages 3842--3852,
  Dublin, Ireland. Association for Computational Linguistics.

\bibitem[{Wu et~al.(2023)Wu, Wang, Wan, Jiao, and Lyu}]{wu2023chatgpt}
Haoran Wu, Wenxuan Wang, Yuxuan Wan, Wenxiang Jiao, and Michael Lyu. 2023.
\newblock \href {http://arxiv.org/abs/2303.13648} {Chatgpt or grammarly?
  evaluating chatgpt on grammatical error correction benchmark}.

\bibitem[{Yasunaga et~al.(2021)Yasunaga, Leskovec, and
  Liang}]{yasunaga-etal-2021-lm}
Michihiro Yasunaga, Jure Leskovec, and Percy Liang. 2021.
\newblock \href {https://aclanthology.org/2021.emnlp-main.611} {{LM}-critic:
  Language models for unsupervised grammatical error correction}.
\newblock In \emph{Proceedings of the 2021 Conference on Empirical Methods in
  Natural Language Processing}, pages 7752--7763, Online and Punta Cana,
  Dominican Republic. Association for Computational Linguistics.

\bibitem[{Zock and Gemechu(2017)}]{zock2017use}
Michael Zock and Debela~Tesfaye Gemechu. 2017.
\newblock Use your mind and learn to write: The problem of producing coherent
  text.
\newblock In \emph{Cognitive Approach to Natural Language Processing}, pages
  129--158. Elsevier.

\end{thebibliography}
\bibliographystyle{acl_natbib}

\clearpage
\appendix

\section{Human Evaluation Experiment Details}

The experiment was carried out using Amazon Mechanical Turk. Participants received compensation at a rate of \$1.7 per HIT, which roughly translated to an hourly wage of \$17. The variation in inter-annotator agreement for scoring five options, as denoted by Cohen's kappa, ranged between 0.41 (for JFLEG) and 0.32 (for BEA-2019).
\end{document}